\documentclass[sigconf]{acmart}

\AtBeginDocument{%
  }

\usepackage{algorithm}
\usepackage{algorithmic}
\usepackage{multirow}

\usepackage{balance}

\usepackage{draftwatermark}  % 这个包用于水印控制
\SetWatermarkText{}  % 设为空字符串，移除水印
\SetWatermarkScale{0}  % 设为 0，完全隐藏水印

\renewcommand\footnotetextcopyrightpermission[1]{}

\AtBeginDocument{\fancyfoot{}} % 确保清除页脚
\begin{document}

%%
%% The "title" command has an optional parameter,
%% allowing the author to define a "short title" to be used in page headers.
\title{Incorporating the Refractory Period into Spiking Neural Networks through Spike-Triggered Threshold Dynamics}

%%
%% The "author" command and its associated commands are used to define
%% the authors and their affiliations.
%% Of note is the shared affiliation of the first two authors, and the
%% "authornote" and "authornotemark" commands
%% used to denote shared contribution to the research.
\author{Yang Li}
\affiliation{%
  \department{School of Computer Science}
  \institution{Sichuan University}
  \city{Chengdu}
  \country{China}
}
\email{liyang0@stu.scu.edu.cn}

\author{Xinyi Zeng}
\affiliation{%
  \department{School of Computer Science}
  \institution{Sichuan University}
  \city{Chengdu}
  \country{China}
}
\email{perperstudy@gmail.com}

\author{Zhe Xue}
\affiliation{%
  \department{School of Computer Science}
  \institution{Sichuan University}
  \city{Chengdu}
  \country{China}
}
\email{zacharyxz@stu.scu.edu.cn}

\author{Pinxian Zeng}
\affiliation{%
\department{School of Computer Science}
 \institution{Sichuan University}
  \city{Chengdu}
  \country{China}
}
\email{zengpinxian@stu.scu.edu.cn}

\author{Zikai Zhang}
\affiliation{%
\department{School of Computer Science}
  \institution{Sichuan University}
  \city{Chengdu}
  \country{China}
}
\email{2024223040071@stu.scu.edu.cn}

\author{Yan Wang}
\authornote{Corresponding author.}
% \authornotemark[1]
\affiliation{%
\department{School of Computer Science}
  \institution{Sichuan University}
  \city{Chengdu}
  \country{China}
}
\email{wangyanscu@hotmail.com}

% \renewcommand{\shortauthors}{Yang Li et al.}
%% No italics, no superscripts
%% Use footnote or author note to identify equal contribution and/or contact author info

%%
%% By default, the full list of authors will be used in the page
%% headers. Often, this list is too long, and will overlap
%% other information printed in the page headers. This command allows
%% the author to define a more concise list
%% of authors' names for this purpose.
% \renewcommand{\shortauthors}{Trovato et al.}

%%
%% The abstract is a short summary of the work to be presented in the
%% article.
\begin{abstract}
  As the third generation of neural networks, spiking neural networks (SNNs) have recently gained widespread attention for their biological plausibility, energy efficiency, and effectiveness in processing neuromorphic datasets. To better emulate biological neurons, various models such as Integrate-and-Fire (IF) and Leaky Integrate-and-Fire (LIF) have been widely adopted in SNNs. However, these neuron models overlook the refractory period, a fundamental characteristic of biological neurons. Research on excitable neurons reveal that after firing, neurons enter a refractory period during which they are temporarily unresponsive to subsequent stimuli. This mechanism is critical for preventing over-excitation and mitigating interference from aberrant signals. Therefore, we propose a simple yet effective method to incorporate the refractory period into spiking LIF neurons through spike-triggered threshold dynamics, termed RPLIF. Our method ensures that each spike accurately encodes neural information, effectively preventing neuron over-excitation under continuous inputs and interference from anomalous inputs. Incorporating the refractory period into LIF neurons is seamless and computationally efficient, enhancing robustness and efficiency while yielding better performance with negligible overhead. To the best of our knowledge, RPLIF achieves state-of-the-art performance on Cifar10-DVS(82.40\%) and N-Caltech101(83.35\%) with fewer timesteps and demonstrates superior performance on DVS128 Gesture(97.22\%) at low latency.
\end{abstract}

%% A "teaser" image appears between the author and affiliation
%% information and the body of the document, and typically spans the
%% page.
% \begin{teaserfigure}
%   \includegraphics[width=\textwidth]{sampleteaser}
%   \caption{Seattle Mariners at Spring Training, 2010.}
%   \Description{Enjoying the baseball game from the third-base
%   seats. Ichiro Suzuki preparing to bat.}
%   \label{fig:teaser}
% \end{teaserfigure}

% \received{20 February 2007}
% \received[revised]{12 March 2009}
% \received[accepted]{5 June 2009}

%%
%% This command processes the author and affiliation and title
%% information and builds the first part of the formatted document.
\maketitle

\section{Introduction}
Over the past few years, artificial neural networks (ANNs) have achieved remarkable success in various domains, such as object recognition ~\cite{redmon2016you}, object segmentation ~\cite{girshick2014rich}, and natural language processing ~\cite{vaswani2017attention}. However, due to their inherent limitation in understanding temporal information, ANNs struggle to handle neuromorphic datasets that contain rich temporal dynamics ~\cite{sironi2018hats,ramesh2019dart}. Spiking neural networks (SNNs), regarded as the next generation of neural networks, naturally excel in processing temporal information, offering high energy efficiency and biological plausibility \cite{roy2019towards}. As a result, SNNs are widely considered the most suitable networks for handling neuromorphic datasets ~\cite{shen2024tim,zhang2025staa}.

Spiking neural networks are inspired by biological neurons in the brain and operate by mimicking the way the brain processes information \cite{maass1997networks}.
Each neuron integrates input currents into its membrane potential and fires a spike only when the potential exceeds the threshold; otherwise, it remains idle. This property makes SNNs highly energy-efficient compared to traditional ANNs. In SNNs, various neuron models have been proposed to better simulate biological properties, such as IF ~\cite{hao2023reducing,xu2023constructing}, LIF \cite{wu2018spatio,yu2025fsta}, and PLIF \cite{fang2021incorporating}. The IF neuron models the fundamental functions of a neuron, including charging, firing, and resetting. Due to its simplicity, many studies use the IF neuron as the foundational neuron model. The LIF neuron is also one of the most widely used neuron models in SNNs ~\cite{zhu2024tcja,li2022neuromorphic}. Compared to the IF neuron, the LIF neuron additionally incorporates the characteristic of membrane permeability in biological neurons, where the membrane potential naturally decays over time. To enhance the learning capability of SNNs, \cite{fang2021incorporating} proposed the PLIF (Parametric Leaky Incorporate-and-Fire) neuron by incorporating learnable membrane time constants, which has also been widely adopted in SNN studies ~\cite{ding2024shrinking,fang2021deep}.

\begin{figure}[tb]
\centering
\includegraphics[
        scale=0.35,        % 或者指定比例
        trim=0 120 0 120, 
        clip
    ]{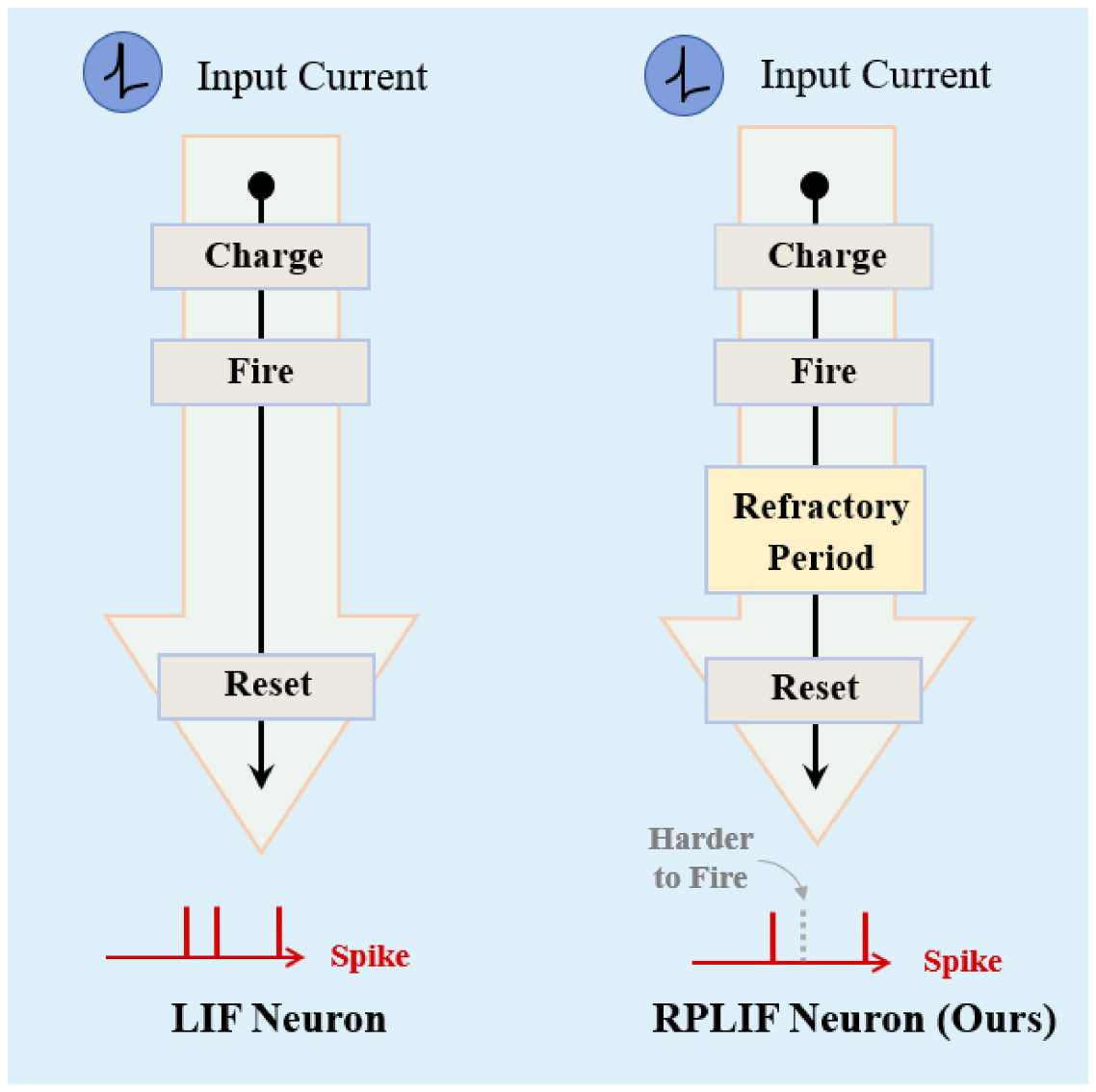}
\caption{The difference between LIF neuron and RPLIF neuron. After a neuron fires a spike, it enters the refractory period in the next time step. During the refractory period, the neuron’s threshold is elevated, making it harder to fire another spike. If no spike is fired during this phase, the neuron returns to its normal state.}
\Description{The two figures illustrate the different processes of RPLIF and LIF during a complete spiking activity cycle and the effect of the refractory period in output spikes}
\label{figure1}
\end{figure}

However, all the aforementioned neuron models do not account for the refractory period, a crucial characteristic of excitable neurons. Biologically, the refractory period functions as both a protective mechanism and a regulatory function in the nervous system, enabling temporal filtering of input signals. Without the refractory period, neurons may generate spikes in response to every input stimulus, potentially firing continuously at extremely high frequencies, leading to neuronal over-excitation. Furthermore, neurons in the absence of the filtering effect of the refractory period might respond to minor noise stimuli, resulting in erroneous spike emissions and distorted signal communication. Such issues, including high-frequency firing and reduced robustness to noise, could significantly degrade the performance of SNNs. Therefore, we advocate for the incorporation of the refractory period into neuronal activity to address these challenges and enhance the robustness of SNNs in object recognition tasks under noisy conditions.

% However, all the aforementioned neuron models do not consider the refractory period, a crucial biological characteristic of excitable neurons. Biologically, the refractory period serves as both a protective mechanism and a regulatory function in the nervous system. Without the constraints of the refractory period, neurons could generate spikes in response to every input stimulus, potentially firing continuously at extremely high frequencies, leading to neuronal over-excitation. For instance, the refractory period enables temporal filtering of input signals. In the absence of the filtering effect of the refractory period, neurons might react to minor noise stimuli, leading to the miscommunication of signals. Such incorrect spike emissions could significantly degrade the performance of spiking neural networks. Therefore, we advocate incorporating the refractory period into neuronal activity to address the potential issues mentioned above and improve the robustness of SNNs against noise in object recognition tasks.

In this work, we present PRLIF, a novel neuron model that incorporates a refractory period phase following the standard firing phase of the vanilla LIF neuron. Compared to IF neuron, the LIF neuron accounts for membrane potential leakage, making it more plausible for incorporating biological features such as the refractory period. Regarding the design of the refractory period, instead of manually setting a fixed refractory period during which neurons cannot spike, we propose a dynamic approach to simulate the refractory period by increasing the threshold at the subsequent timestep after firing a spike. This method allows us to model the refractory period with minimal computational overhead. The distinction between our PRLIF model and the standard LIF neuron is illustrated in Fig.~\ref{figure1}. Our experimental results demonstrate that RPLIF not only enhances the learning capabilities of SNNs but also improves their robustness to noise.

% In this work, we introduce the integrated LIF neuron, referred to as PRLIF, by introducing a refractory period phase after the vanilla firing phase. We use the LIF neuron as the foundational model because, compared to the IF neuron, the LIF neuron accounts for membrane potential leakage, making it more biologically plausible. We believe the LIF neuron is better suited to incorporate the biological feature of the refractory period. Instead of manually setting a fixed period during which neurons cannot spike, we simulate the refractory period by increasing the threshold at the subsequent timestep after firing a spike. In this way, we can simulate the refractory period with almost no additional computational cost. The difference between our RPLIF and the vanilla LIF is illustrated in Fig.~\ref{figure1}. Our experiments show that incorporating the refractory period into spiking neurons enhances the learning of SNNs and their robustness to noise.

Our main contributions can be summarized as follows:

\begin{itemize}
    \item We identify that current SNN neurons overlook the biological phenomenon of the refractory period, making them overly sensitive to continuous and noisy inputs. To address this, we propose a spike-triggered threshold dynamics method to simulate the refractory period, effectively leveraging input information and reducing computational overhead. 
    % \item We adopt a spike-triggered threshold dynamics method to simulate the refractory period. This approach can better leverage input information, especially in light of the current trend toward reducing timesteps in SNNs. Moreover, it effectively models the refractory period with minimal computational overhead.
    \item To better incorporate the refractory period into modern low-timestep SNNs, we decouple the biologically continuous absolute and relative refractory periods, treating them as two distinct implementations of RPLIF. This separation ensures that neurons can generate a sufficient number of spikes to effectively transmit information forward.
    % \item We conducted extensive experiments on three neuromorphic datasets—CIFAR10-DVS \cite{li2017cifar10}, N-Caltech101 \cite{orchard2015converting}, and DVS128 \cite{amir2017low}—to evaluate the effectiveness of RPLIF. With fewer timesteps, RPLIF achieves state-of-the-art accuracy of 82.40\% on CIFAR10-DVS and 83.35\% on N-Caltech101. Additionally, it outperforms existing methods on DVS128 under very low latency, achieving an impressive accuracy of 97.22\%. Additionally, extensive exploratory experiments were conducted to validate the robustness and efficiency of RPLIF.
    \item Experimental results show that RPLIF outperforms previous methods on neuromorphic datasets for classification tasks under very low latency. Additional exploratory experiments further validate its robustness and efficiency.
\end{itemize}

\section{Related Work}
\subsection{SNN Learning Methods}

Currently, there are two main approaches to achieving high performance SNNs. 
% The first approach involves converting ANNs into SNNs by mapping the parameters of ANNs onto corresponding SNN architectures and replacing ReLU with a spiking neuron ~\cite{bu2023optimal}. This enables the converted SNNs to retain some of the knowledge of the original ANN structures. Such converted SNNs bypass the challenge of directly training SNNs, where the gradient cannot propagate backward effectively. Moreover, these SNNs require only minimal fine-tuning instead of extensive training to achieve satisfactory performance. Many recent studies focus on how to efficiently convert ANNs into high-performance SNNs. However, this method has inherent limitations. Since the parameters of the SNN are derived from the ANN, the accuracy of the converted SNN is inherently constrained and often cannot surpass that of the original ANN. Furthermore, as ANNs inherently cannot handle neuromorphic datasets well, the converted SNNs also fail to learn the temporal relationships in data. These two factors together limit the performance of this approach.
The first approach converts ANNs to SNNs by mapping ANN parameters to SNNs and replacing ReLU with spiking neurons ~\cite{bu2023optimal}, allowing the SNN to inherit some knowledge from the ANN. This avoids the difficulty of training SNNs directly where the gradient cannot propagate backward effectively and requires only minimal fine-tuning. However, since the parameters of the SNN are derived from the ANN, its performance often cannot exceed the original ANN's accuracy.
Furthermore, as ANNs inherently cannot handle neuromorphic datasets well, the converted SNNs also fail to learn the temporal relationships in data. These two factors together limit the performance of this approach.
% and struggles with temporal learning, making it unsuitable for neuromorphic datasets.

In recent years, directly training SNNs has gained increasing popularity ~\cite{zheng2021going,shen2024tim}. To address the challenge of non-propagatable gradients, \cite{neftci2019surrogate} have proposed the surrogate gradient method, which retains Heaviside step function during forward propagation, while employing a surrogate function for backpropagation. By doing so, the problem of directly training SNNs is effectively resolved. \cite{zheng2021going} introduced a novel threshold dependent batch normalization method to mitigate the challenges of vanishing and exploding gradients.
\cite{wang2023adaptive} proposed to gradually evolve a prototype neural network into an SNN by incorporating learnable relaxation degrees and random spike noise, mitigating the gradient mismatching problem. \cite{ma2023exploiting} introduced noisy SNNs and a noise-driven learning rule by modeling stochastic neuronal dynamics, enhancing robustness over deterministic SNNs under challenging perturbations.
~\cite{xu2024reversing,xu2023constructing} proposed training SNNs via knowledge distillation and ~\cite{chowdhury2021spatio,li2024towards} introduced a pruning-based approach, both offering new perspectives for achieving high-performance and efficient SNNs.
% ~\cite{xu2024reversing,xu2023constructing} proposed a knowledge distillation-based training paradigm for SNNs, offering a novel perspective for directly training SNNs.
Compared to the approach of converting ANNs to SNNs, training SNNs with surrogate gradients offers several advantages. Firstly, directly trained SNNs can process neuromorphic datasets and are well-suited for deployment on neuromorphic chips such as Loihi \cite{davies2018loihi} and Truenorth \cite{akopyan2015truenorth}. Additionally, the conversion-based approach often requires thousands of timesteps to sufficiently capture the information of the ANN, resulting in significant time costs. In contrast, directly trained SNNs can achieve results comparable to, or even surpass, ANNs with few timesteps.
% Compared to ANN-to-SNN conversion, training SNNs with surrogate gradients offers key benefits. It enables direct processing of neuromorphic datasets and efficient deployment on chips like Loihi\cite{davies2018loihi} and TrueNorth\cite{akopyan2015truenorth}. While conversion methods often require thousands of timesteps, leading to high time costs, directly trained SNNs can match or exceed ANN performance with far fewer timesteps.
The method adopted in this paper falls into this category.

\subsection{Firing Threshold Plasticity}

Firing Threshold Plasticity is a fundamental yet crucial mechanism in neurons, determining their sensitivity to input signals and regulating their firing rate. The selection of a firing threshold is particularly significant: if the threshold is set too low, differences between input signals may be overlooked, as neurons will generate a fixed spike whenever the input exceeds the threshold, failing to fully utilize the information contained in the inputs. Conversely, if the threshold is set too high, neurons will struggle to fire spikes, hindering the transmission of useful information forward. Therefore, the appropriate selection and adjustment of firing threshold have become critical research areas in recent years. 

Several studies have explored innovative approaches to optimize threshold mechanisms. DIET-SNN \cite{rathi2021diet} proposed training the weights of the network along with the membrane leak and firing threshold of LIF neurons during backpropagation to optimize both accuracy and latency. STL-SNN \cite{sun2023synapse} introduced a synaptic-threshold co-learning method for directly training SNNs. This collaborative learning not only enables SNNs to perform competitively across datasets with various tasks but also endows them with strong robustness, stable signal transmission, and efficient energy consumption. MT-SNN \cite{wang2023mt} mitigates the inherent accuracy loss of SNNs compared to floating-point neural networks by adopting multiple thresholds. In \cite{chen2022adaptive}, an adaptive threshold mechanism is proposed, allowing thresholds to dynamically adjust based on input data. This approach achieves thresholds that are sufficiently low to differentiate inputs while maintaining discrimination capability. Ternary Spike \cite{guo2024ternary} employs both positive and negative thresholds, greatly increasing the information capacity of neurons.

\begin{figure*}[tb]
\centering
\includegraphics[
        width=\textwidth,        % 或者指定比例
        trim=0 335 0 350, 
        clip
    ]{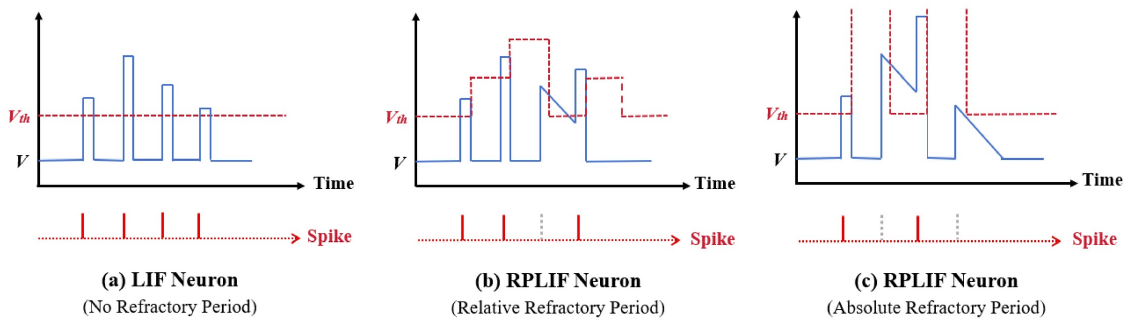}
\caption{Vanilla LIF neurons (a), RPLIF neurons with relative refractory period (b), and RPLIF neurons with absolute refractory period (c). When RPLIF is in the relative refractory period, the threshold is higher, but if the input current is strong enough, it can still generate spikes. After a spike is fired, the threshold continues to increase. When RPLIF is in the absolute refractory period, the threshold reaches a very high value, and in this phase, spikes cannot be generated. The gray line represents spikes that are suppressed due to the increase in the threshold.}
\Description{The three figures illustrate the dynamic changes in membrane potential and threshold for LIF, RPLIF (relative), and RPLIF (absolute) under the same input current.}
\label{figure2}
\end{figure*}

However, most of these adaptive threshold adjustments are based solely on training through backpropagation. Once the training phase is completed, the threshold remains fixed and cannot dynamically change over time. This oversight renders neurons overly sensitive to continuous and anomalous inputs, thereby limiting their ability to analyze continuous temporal dynamics and anomalous changes in neuromorphic data. The issue is particularly pronounced in SNNs when handling sparse datasets such as neuromorphic datasets, where incorrect spike firing can significantly degrade the performance of SNNs. \cite{chen2022adaptive} proposed adjusting the threshold based on input data, whereas in our approach, the threshold is dynamically adjusted based on output data, i.e., spikes, considering the impact that the act of firing spikes should have on the threshold. To this end, in this paper, we propose a simple yet effective method of simulating the refractory period by adjusting the threshold for the next time step after a neuron fires a spike, which enhances the neuron’s robustness to continuous and anomalous inputs.

\section{Preliminary}

% \subsection{LIF Model}

Unlike ANNs, SNNs rely on binary spikes to transmit information between layers. Each neuron receives an input current and accumulates it into the membrane potential. When the membrane potential exceeds a predefined threshold, the neuron fires a spike; otherwise, it remains idle. Previous studies have proposed various neuron models with different membrane potential dynamics to simulate the functions of biological neurons. In this work, we adopt the widely used LIF neuron, which achieves a reasonable balance between simplicity and biological plausibility. Without decaying the input and setting $V_{\text{reset}}$ to 0, the membrane potential dynamics of an LIF neuron can be described as follows:
\begin{equation}
    U[t] = (1 - \frac{1}{\tau})V[t-1] + I[t]
    \label{eq:eq1}
\end{equation}%

\begin{equation}
    S[t] = \Theta(U[t] - V_{\text{th}})
    \label{eq:spike}
\end{equation}

\begin{equation}
    V[t] = U[t](1 - S[t])
\end{equation}

\noindent where \( I[t] \) represents the input current at time step \( t \), \( U[t] \) denotes the membrane potential after receiving the input, \( S[t] \) indicates whether a spike is fired, \( V_{\text{th}} \) represents the threshold, \( V[t] \) is the membrane potential after spike firing phase, \(\tau\) is a time constant, and \( \Theta[x] \) denotes the Heaviside step function. We use the arctangent function proposed in \cite{fang2021incorporating} as the surrogate function, represented by Eq.~\eqref{eq:derivative}:
\begin{equation}
    \frac{\partial S[t]}{\partial U[t]} = \frac{1}{1 + x^2}
    \label{eq:derivative}
\end{equation}

\begin{figure}[tb]
\centering

\includegraphics[
        width=0.48\textwidth,        % 或者指定比例
        trim=0 220 0 200, 
        clip
    ]{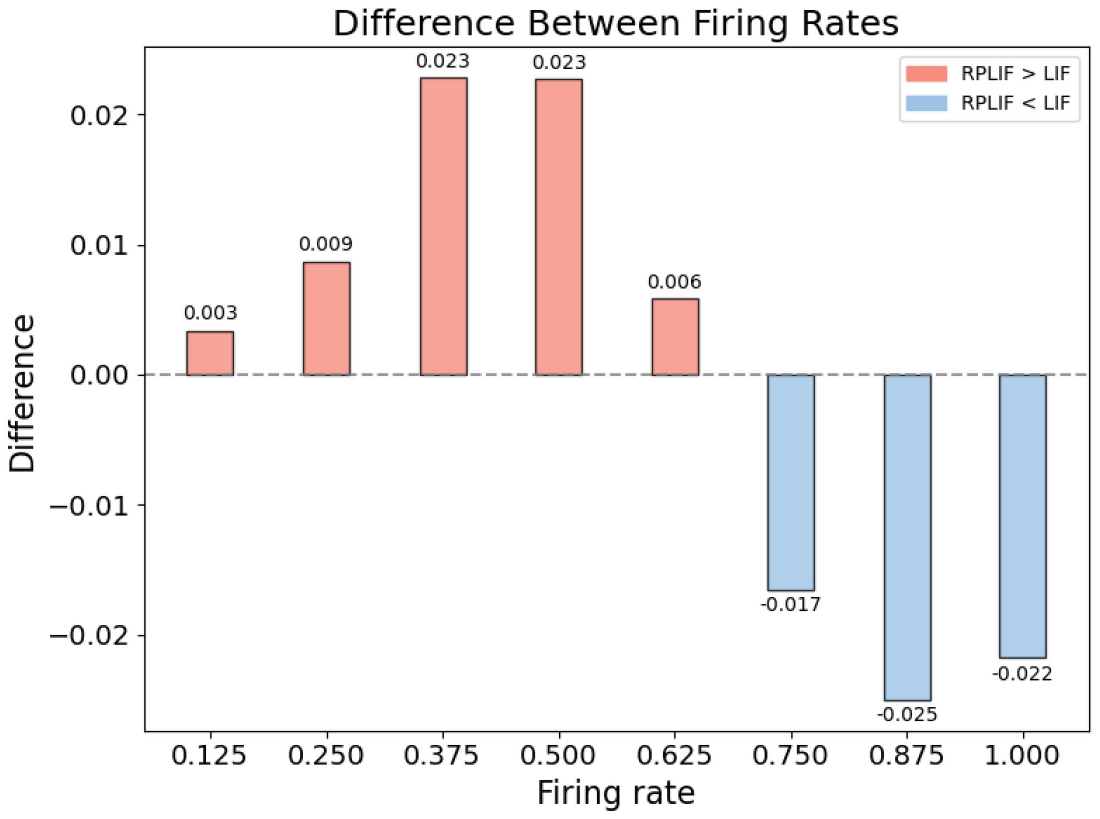}
\caption{The difference between firing rates of RLIF and LIF with 8 timesteps. LIF neurons exhibit a significantly higher proportion of firing rates close to 1 compared to RPLIF neurons. This phenomenon can be attributed to the susceptibility of LIF neurons to continuous and abnormal inputs. In contrast, RPLIF maintains the firing rate within a more reasonable range, thereby enhancing the robustness of SNNs.}
\Description{The bar chart visualizes the difference in firing rates between RPLIF and LIF neurons. The red bars represent the firing rate of RPLIF, while the blue bars represent that of LIF.}
\label{figure_3}
\end{figure}

\section{Proposed Method}

\subsection{Motivation and Refractory Period}

Current research on neuron thresholds mainly focuses on selecting more suitable initial thresholds and dynamically co-adjusting thresholds and weights to achieve more reasonable firing rates. However, these methods often overlook the biological principles underlying the threshold mechanism. In 1952, Alan Hodgkin and Andrew Huxley published a series of seminal papers \cite{hodgkin1952quantitative} that provided a comprehensive explanation of the dynamic processes involving sodium and potassium ion channels in the generation of spikes. Building upon this foundational work, the concept of the refractory period was gradually developed.

In brief, after an excitable neuron responds to a stimulus, there exists a period during which it does not respond to any subsequent stimuli, regardless of their intensity. This period is known as the refractory period, which can be further divided into two phases. The first is called absolute refractory period. Following the first effective stimulus, the excitability of the neuron drops to zero for a short duration, and the threshold becomes infinite. During this phase, no matter how strong the subsequent stimulus is, it cannot elicit another spike. After the absolute refractory period, the membrane potential remains at a lower level (close to or below the resting potential), named relative refractory period. During this phase, a stronger-than-usual stimulus is required to bring the membrane potential to the threshold and trigger a new spike. The refractory period plays a crucial role in regulating firing frequency, effectively preventing excessive neuronal excitation and ensuring stable neural activity. Both the absolute and relative refractory periods reflect the dynamic changes in excitability experienced by neurons during a single excitation. These phases significantly influence how excitable neurons respond to subsequent stimuli, especially when subjected to continuous stimulation. Thus, incorporating this biological characteristic into spiking neurons is essential.

Inspired by this, we propose a straightforward yet elegant method closely aligned with the inherent dynamics of neurons by adjusting the threshold of a neuron at the time step immediately following a spike to effectively simulate the refractory period. This mechanism promotes temporal selectivity, enabling the network to respond more discriminatively to salient input features over time. By reducing redundant or spurious activations, the refractory effect encourages more compact and distinctive representations in the high-dimensional feature space, ultimately improving the separability of different input patterns. In Fig.~\ref{figure_3}, We analyzed the firing rate distribution in the first Conv with 8 timesteps. Our method aligns closely with the intrinsic dynamics of neurons and effectively addresses the issue of excessive excitation under continuous inputs, leading to a more proper firing rate.

\subsection{Spike-Triggered Threshold Dynamics}

Several works ~\cite{ikegawa2022rethinking,ma2018novel} have attempted to incorporate the refractory period into neuron activity before by suppressing neuron firing by resetting the membrane potential after a spike. However, our approach differs from these methods in the following aspects. First, we are the first to incorporate the refractory period into existing neurons in directly trained SNNs using surrogate gradients while previous methods primarily focus on ANNs to SNNs conversion methods. Second, previous approaches typically simulate the absolute refractory period by forcing the membrane potential to remain at zero during a manually set period of time. When input currents enter the neuron during such period, all information carried by these inputs is completely discarded. In ANN to SNN conversion methods, which typically require dozens to hundreds of time steps, it is still acceptable to simulate the refractory period even at the cost of losing information from some time steps. However, it hardly meets the requirements of low time steps(fewer than twenty time steps) in current directly trained SNNs and fails to make full use of the input information. Given the current trend of reducing the number of time steps, most directly trained SNNs adopt fewer than ten time steps. If the input of the subsequent time steps is discarded after each spike, the SNN may not have sufficient information to complete the recognition task. Therefore, We argue that the input current should instead be retained and decay over time rather than being entirely discarded, which is a more biologically plausible and effective way to incorporate the refractory period into the spiking neurons of current SNNs. To achieve this, we propose a more integrative approach for spiking neurons. Instead of fixing the membrane potential to zero during the absolute refractory period, we set the threshold to a sufficiently large value. This ensures that no spikes are fired during this phase while still preserving and utilizing the input information. 

\begin{algorithm}[tb]
    \caption{Training Framework of RPLIF}
    \label{alg:rplif_training}
    \textbf{Require}: Input \(\mathbf{I}\), Timesteps \(\mathbf{T}\), Initial Threshold \(\mathbf{V_{\text{init\_th}}}\)\\
    \begin{algorithmic}[1] %[1] enables line numbers
        \STATE \(\mathbf{V_{\text{th}}} \gets \mathbf{V_{\text{init\_th}}}\) // Initialize the \( V_{\text{th}} \) with \( V_{\text{init\_th}} \)
        \FOR{\( t = 0 \) to \( \mathbf{T - 1} \) }
            \STATE \(\mathbf{V} \gets \mathbf{I}\) // Charge according to Eq.~(\ref{eq:eq1}).
            \STATE \(\textbf{spike} \gets \mathbf{\Theta}(\mathbf{V} - \mathbf{V_{\text{th}}})\) // Fire spikes according to Eq.~(\ref{eq:spike}).
            \STATE \(\mathbf{V} \gets \mathbf{V} \cdot (1 - \textbf{spike})\) // Reset the membrane potential
            \IF{\(\textbf{spike} = 1\)}
                \STATE \(\mathbf{V_{\text{th}}} \gets \mathbf{V_{\text{th}}} \cdot \mathbf{\alpha}\) // Increase the threshold based on the refractory period.
            \ENDIF
            \IF{\(\textbf{spike} = 0\) \textbf{and} \(\mathbf{V_{\text{th}}} > \mathbf{V_{\text{init\_th}}}\)}
                \STATE \(\mathbf{V_{\text{th}}} \gets \mathbf{V_{\text{init\_th}}}\) // Reset \( V_{\text{th}} \) to \( V_{\text{init\_th}} \)
            \ENDIF
        \ENDFOR
    \end{algorithmic}
\end{algorithm}

Furthermore, prior works have primarily focused on modeling only the absolute refractory period, during which neurons are unable to fire for a pre-defined duration. However, we believe that this approach has limitations, particularly when processing neuromorphic datasets. In these datasets, the mainstream approach is to aggregate event sequences into frames for processing. This frame-based compression already spans multiple event lengths. If only the absolute refractory period is considered—i.e., prohibiting spiking for a fixed number of timesteps after firing—SNNs may lose some temporal information, resulting in suboptimal performance. In the meantime, the relative refractory period is a more balanced approach, during which sufficiently strong stimuli can still trigger neuronal spikes. This is more conducive to information transfer at low time steps. Therefore, we propose decoupling the biologically continuous absolute and relative refractory periods, treating them as two distinct mechanisms to conduct experiments.

Building on the above analysis, we propose a simple yet effective method to incorporate the refractory period into spiking neurons. After a neuron fires a spike, the threshold for the subsequent timestep is multiplied by \( \alpha \), which is set as a pre-defined hyperparameter. To simulate the relative refractory period, we set \( \alpha \) between 1 and 2. Conversely, to simulate the absolute refractory period, \( \alpha \) is assigned a sufficiently large value, making sure no spikes will be fired during this period, and in our work, we set it to 100. During refractory period, the neuron continues to receive input currents, which decay over time. In Fig.~\ref{figure2}, we illustrate the evolution of the neuron’s membrane potential and threshold as they respond to input currents over a period of time.

\subsection{RPLIF Neuron}

To better emulate biological neurons, this study adopts the LIF neuron as the foundational neuron model. In Section 3, we have introduced the dynamics equations of LIF neuron. To incorporate the refractory period into the LIF neuron, modifications to certain equations are required. After firing a spike, i.e., following Eq.~\eqref{eq:spike}, the threshold needs to be updated as follows:
\begin{equation}
    V_{\text{th}} = V_{\text{init\_th}}(1 - S[t]) + \alpha V_{\text{th}} S[t]
\end{equation}

\noindent where \( V_{\text{init\_th}} \) represents the initial threshold, and \( \alpha \) is a parameter that controls the increase in the threshold. We name the neuron as RPLIF. In Algorithm \ref{alg:rplif_training}, we provide the complete pseudocode for the RPLIF neuron.

% \begin{algorithm}[tb]
%     \caption{Training Framework of RPLIF}
%     \label{alg:rplif_training}
%     \textbf{Require}: Input \(\mathbf{I}\), Timesteps \(\mathbf{T}\), Initial Threshold \(\mathbf{V_{\text{init\_th}}}\)\\
%     \begin{algorithmic}[1] %[1] enables line numbers
%         \STATE \(\mathbf{V_{\text{th}}} \gets \mathbf{V_{\text{init\_th}}}\) // Initialize the \( V_{\text{th}} \) with \( V_{\text{init\_th}} \)
%         \FOR{\( t = 0 \) to \( \mathbf{T - 1} \) }
%             \STATE \(\mathbf{V} \gets \mathbf{I}\) // Charge according to Eq.~(\ref{eq:eq1}).
%             \STATE \(\textbf{spike} \gets \mathbf{\Theta}(\mathbf{V} - \mathbf{V_{\text{th}}})\) // Fire spikes according to Eq.~(\ref{eq:spike}).
%             \STATE \(\mathbf{V} \gets \mathbf{V} \cdot (1 - \textbf{spike})\) // Reset the membrane potential
%             \IF{\(\textbf{spike} = 1\)}
%                 \STATE \(\mathbf{V_{\text{th}}} \gets \mathbf{V_{\text{th}}} \cdot \mathbf{\alpha}\) // Increase the threshold based on the refractory period.
%             \ENDIF
%             \IF{\(\textbf{spike} = 0\) \textbf{and} \(\mathbf{V_{\text{th}}} > \mathbf{V_{\text{init\_th}}}\)}
%                 \STATE \(\mathbf{V_{\text{th}}} \gets \mathbf{V_{\text{init\_th}}}\) // Reset \( V_{\text{th}} \) to \( V_{\text{init\_th}} \)
%             \ENDIF
%         \ENDFOR
%     \end{algorithmic}
% \end{algorithm}

\begin{table}
    \centering
    \caption{The top-1 accuracy of CIFAR10-DVS with different \( \alpha \) values. The baseline, RPLIF (Relative), and RPLIF (Absolute) are aligned with the three neurons shown in Fig.~\ref{figure2}.}
    \begin{tabular}{llc}
        \toprule
        Model & $\alpha$ & Accuracy (\%) \\
        \midrule
        Baseline & 1.0  & 76.77 \\
        \midrule
        \multirow{3}{*}{RPLIF (Relative)} 
        & 1.4 & 78.02 \\
        & 1.5 & \textbf{78.96} \\
        & 1.6 & 78.54 \\
        \midrule
        RPLIF (Absolute) & 100 & 78.33 \\
        \bottomrule
    \end{tabular}
    \label{tab:alpha_accuracy}
\end{table}

\section{Experimental Results}

\subsection{Datasets and Implementation Details}

To evaluate the effectiveness of RPLIF, we conducted extensive experiments on three neuromorphic datasets: CIFAR10-DVS \cite{li2017cifar10}, N-CALTECH101 \cite{orchard2015converting}, and DVS128 Gesture \cite{amir2017low}. 
CIFAR10-DVS is an event-driven version of CIFAR-10 generated using a Dynamic Vision Sensor, including 10k images. N-CALTECH101 is the dynamic vision variant of the Caltech101 dataset, covering 101 object categories. DVS128 Gesture, based on the DVS128 sensor, captures dynamic human gestures through event streams, offering 11 gesture categories under varying light conditions.
It is important to note that for static datasets, the common practice involves replicating a single static image across \(T \) timesteps. This creates a fixed input sequence without any temporal variation, thereby introducing minimal temporal redundancy. Therefore, to ensure the validity and relevance of our experiments, we specifically tailored our analysis for neuromorphic datasets to investigate the impact of the refractory mechanism on time-dependent data.

\begin{table*}
    \centering
    \caption{Performance comparison with SOTA methods on CIFAR10-DVS dataset. $^{*}$ indicates results with data augmentation.}
    \begin{tabular}{clclcc}
        % \hline
        \toprule[1.0pt]
        \multirow{2}{*}{\textbf{Dataset}} & \multirow{2}{*}{\textbf{Method}} & \multirow{2}{*}{\textbf{Venue}} & \multirow{2}{*}{\textbf{Network}} & \multirow{2}{*}{\textbf{Timesteps}} & \multirow{2}{*}{\textbf{Accuracy (\%)}} \\
        & & & & \\
        % \hline
        \toprule[1.0pt]
        % \midrule
        \multirow{14}{*}{CIFAR10-DVS} & Rollout \cite{kugele2020efficient} & Front. Neurosci. ’2020 & VGG-16 & 48 & 65.61 \\
        % & Rollout \cite{kugele2020efficient} & VGG-16 & 48 & 65.61 \\
        & SALT \cite{kim2021optimizing} & Neural Networks ’2019 & VGG-11 & 20 & 67.10 \\
        & STBP-tdBN \cite{zheng2021going} & AAAI ’2021 & ResNet-19 & 10 & 67.80 \\
        & LIAF-Net \cite{wu2021liaf} & IEEE Trans. ’2021 & VGG-like & 10 & 70.40 \\
        % & SALT \cite{kim2021optimizing} & VGG-11 & 20 & 67.10 \\
        % & Rollout \cite{kugele2020efficient} & VGG-16 & 48 & 65.61 \\
        & Dspike \cite{li2021differentiable} & NeurIPS ’2021 & ResNet-18$^{*}$ & 10 & 75.40 \\
        & DSR \cite{meng2022training} & CVPR ’2022 & VGG-11 & 20 & 77.27 \\
        & NDA \cite{li2022neuromorphic} & ECCV ’2022 & VGG-11$^{*}$ & 10 & 79.60 \\
        & TET \cite{deng2022temporal} & ICLR ’2022 & VGG-9 & 10 & 77.33 \\
        & SLTT \cite{meng2023towards} & ICCV ’2023 & VGG-11 & 10 & 77.17 \\
        % & SLSSNN (Xu et al. 2023) & VGG-16 & 8 & 77.5 \\
        & SSNN \cite{ding2024shrinking} & AAAI ’2024 & VGG-9 & 8 & 78.57 \\
        & TCJA-SNN \cite{zhu2024tcja} & TNNLS ’2024 & VGG-9$^{*}$ & 10 & 80.70 \\
        \cline{2-5}
        & \multirow{2}{*}{\textbf{RPLIF (Ours)}} & - & VGG-9 & \textbf{8} & \textbf{78.96} \\
        & & - & VGG-9$^{*}$ & \textbf{8} & \textbf{82.40} \\
        % \hline
        \bottomrule[1.0pt]
    \end{tabular}
    \label{cifar10dvs}
\end{table*}

\begin{table*}[t]
    \centering
    \caption{Performance comparison with SOTA methods on N-Caltech101 dataset. $^{*}$ indicates results with data augmentation.}
    \begin{tabular}{clclcc}
        \toprule[1.0pt]
        \multirow{2}{*}{\textbf{Dataset}} & \multirow{2}{*}{\textbf{Method}} & \multirow{2}{*}{\textbf{Venue}} & \multirow{2}{*}{\textbf{Network}} & \multirow{2}{*}{\textbf{Timesteps}} & \multirow{2}{*}{\textbf{Accuracy (\%)}} \\
        & & & & \\
        \toprule[1.0pt]
        \multirow{12}{*}{N-CALTECH101} & HATS \cite{sironi2018hats} & CVPR ’2018 & N/A & N/A & 64.20 \\
        & DART \cite{ramesh2019dart} & IEEE Trans. ’2019 & N/A & N/A & 66.42 \\
        & SALT \cite{kim2021optimizing} & Neural Networks ’2019 & VGG-11 & 20 & 55.00 \\
        % & EventMix \cite{shen2023eventmix} & ResNet-18$^{*}$ & 10 & 79.47 \\
        & NDA \cite{li2022neuromorphic} & ECCV ’2022 & VGG-9$^{*}$ & 10 & 78.20 \\
        & tdBN+NDA \cite{li2022neuromorphic} & ECCV ’2022 & ResNet-19$^{*}$ & 10 & 78.60 \\
        % & TCJA-SNN \cite{zhu2024tcja} & VGG-9 & 14 & 78.5 \\
        & EventMix \cite{shen2023eventmix} & Inf. Sci ’2023 & ResNet-18$^{*}$ & 10 & 79.47 \\
        & STCA-SNN \cite{wu2023stca} & Front. Neurosci. ’2023 & VGG-7 & 14 & 80.88 \\
        % & SSNN \cite{ding2024shrinking} & VGG-9 & 8 & 79.25 \\
        % & Event Transformer \cite{jiang2022event} & Transformer & N/A & 78.9 \\
        & RM SNN \cite{yao2023sparser} & Neural Networks ’2023 & PLIF-SNN & 10 & 81.20 \\
        & TIM \cite{shen2024tim} & IJCAI ’2024 & Spikformer & 10 & 79.00 \\
        & TCJA-SNN \cite{zhu2024tcja} & TNNLS ’2024 & VGG-9 & 14 & 78.50 \\
        & SSNN \cite{ding2024shrinking} & AAAI ’2024 & VGG-9 & 8 & 79.25 \\
        \cline{2-5}
        & \textbf{RPLIF (Ours)} & - & VGG-9 & \textbf{8} & \textbf{83.35} \\
        \bottomrule[1.0pt]
    \end{tabular}
    \label{ncaltech101}
\end{table*}

In the following sections, we conduct extensive ablation and exploratory experiments, as well as compare our method with previous SOTA approaches to validate the effectiveness of RPLIF. All the experiments were conducted on four V100S GPUs, and all implementations were carried out using the SpikingJelly\cite{fang2023spikingjelly} framework, which is a deep learning framework designed for SNNs building on PyTorch. Preprocessing for all three datasets was performed following the methods provided by SpikingJelly. For CIFAR10-DVS and N-CALTECH101, the training and testing sets were split in a 9:1 ratio, while DVS128 Gesture used its predefined training and testing splits. In our experiments, we set the batch size to 16 and the number of epochs to 500. The initial learning rate was set to 0.001 with a cosine decay strategy, and we used the Adam optimizer. Except for CIFAR10-DVS, we set the batch size to 64 and the number of epochs to 250. The initial threshold of RPLIF neurons is set to 1 and all experiments employed automatic mixed precision (AMP), which reduces memory consumption and accelerates training. However, AMP may lead to slight accuracy degradation. During the reset phase, we followed ~\cite{zenke2021remarkable,fang2021incorporating} by separating the reset phase from the computational graph which enhances SNN performance.

\subsection{Ablation Study}

We explored the influence of \( \alpha \) under five different settings: 1, 1.4, 1.5, 1.6, and 100. When \( \alpha = 1 \), the model utilizes standard LIF neurons, serving as the baseline. For \( \alpha \) values of 1.4, 1.5, and 1.6, the model incorporates relative refractory periods with the degree of threshold enhancement increasing proportionally with \( \alpha \). When \( \alpha = 100 \), absolute refractory periods are introduced, during which neurons are completely inhibited from firing.  

Table \ref{tab:alpha_accuracy} presents the experimental results under these five \( \alpha \) values. The results clearly demonstrate that RPLIF neurons consistently outperform standard LIF neurons. Notably, when \( \alpha = 1.5 \), RPLIF achieves the best trade-off between suppressing neuron over-excitation and maintaining effective information transmission, yielding an improvement of over 2\% compared to the baseline. Even when \( \alpha = 100 \), incorporating absolute refractory periods, the model still surpasses the performance of standard LIF neurons. These findings underscore the effectiveness of considering refractory periods in neural modeling. Based on these results, all subsequent experiments are conducted under the \( \alpha = 1.5 \) setting.

% \begin{table*}[h]
%     \centering
%     \caption{Performance comparison with SoTA methods on DVS128 dataset. Use $^{\dag}$ to indicate results reproduced with open-source code}
%     \begin{tabular}{cllcc}
%         \toprule[1.0pt]
%         \multirow{2}{*}{\textbf{Dataset}} & \multirow{2}{*}{\textbf{Method}} & \multirow{2}{*}{\textbf{Network}} & \multirow{2}{*}{\textbf{Timesteps}} & \multirow{2}{*}{\textbf{Accuracy (\%)}} \\
%         & & & & \\
%         \toprule[1.0pt]
%         \multirow{8}{*}{DVS128} & SLAYER \cite{shrestha2018slayer} & SNN (8 layers) & 1600 & 93.64 \\
%         & DECOLLE \cite{kaiser2020synaptic} & SNN (3 layers) & 500 & 95.54 \\
%         & tdBN \cite{zheng2021going} & PLIF-Net$^{\dag}$ & 8 & 93.75 \\
%         & TEBN \cite{duan2022temporal} & PLIF-Net$^{\dag}$ & 8 & 93.06 \\
%         & PLIF \cite{fang2021incorporating} & PLIF-Net$^{\dag}$ & 8 & 94.79 \\
%         & TCJA-SNN \cite{zhu2024tcja} & PLIF-Net$^{\dag}$ & 8 & 94.44 \\
%         & SSNN \cite{ding2024shrinking} & VGG-9 & 8 & 94.91 \\
%         \cline{2-5}
%         & \multirow{2}{*}{\textbf{RPLIF (Ours)}} & PLIF-Net & \textbf{8} & \textbf{95.49} \\
%         & & VGG-9 & \textbf{8} & \textbf{97.22} \\
%         \bottomrule[1.0pt]
%     \end{tabular}
%     \label{dvs128}
% \end{table*}

\begin{table*}[h]
    \centering
    \caption{Performance comparison with SoTA methods on DVS128 dataset. $^{\dag}$ indicates results reproduced with open-source code.}
    \begin{tabular}{clclcc}
        \toprule[1.0pt]
        \multirow{2}{*}{\textbf{Dataset}} & \multirow{2}{*}{\textbf{Method}} & \multirow{2}{*}{\textbf{Venue}} & \multirow{2}{*}{\textbf{Network}} & \multirow{2}{*}{\textbf{Timesteps}} & \multirow{2}{*}{\textbf{Accuracy (\%)}} \\
        & & & & \\
        \toprule[1.0pt]
        \multirow{8}{*}{DVS128} & SLAYER \cite{shrestha2018slayer} & NeurIPS ’2018 & SNN (8 layers) & 1600 & 93.64 \\
        & DECOLLE \cite{kaiser2020synaptic} & Front. Neurosci ’2020 & SNN (3 layers) & 500 & 95.54 \\
        & tdBN \cite{zheng2021going} & AAAI ’2021 & PLIF-Net$^{\dag}$ & 8 & 93.75 \\
        & TEBN \cite{duan2022temporal} & NeurIPS ’2022 & PLIF-Net$^{\dag}$ & 8 & 93.06 \\
        & PLIF \cite{fang2021incorporating} & ICCV ’2021 & PLIF-Net$^{\dag}$ & 8 & 94.79 \\
        & TCJA-SNN \cite{zhu2024tcja} & TNNLS ’2024 & PLIF-Net$^{\dag}$ & 8 & 94.44 \\
        & SSNN \cite{ding2024shrinking} & AAAI ’2024 & VGG-9 & 8 & 94.91 \\
        \cline{2-5}
        & \multirow{2}{*}{\textbf{RPLIF (Ours)}} & - & PLIF-Net & \textbf{8} & \textbf{95.49} \\
        & & - & VGG-9 & \textbf{8} & \textbf{97.22} \\
        \bottomrule[1.0pt]
    \end{tabular}
    \label{dvs128}
\end{table*}

\subsection{Comparison with SOTA Methods}

\textbf{CIFAR10-DVS:} Since the original resolution is 128×128, we downsampled it to 48×48, following \cite{zhu2024tcja}, to reduce computational costs and memory consumption, and we adopted the same VGG9-like network architecture to conduct our experiments. Our experiments were carried out under two conditions: with and without data augmentation. For data augmentation, we employed random horizontal flipping and rolling as in \cite{zhu2024tcja}. As Table \ref{cifar10dvs} shows, the experimental results clearly demonstrate that the RPLIF method outperforms all previous approaches, achieving superior accuracy with fewer timesteps. Without data augmentation, the best accuracy achieved by previous methods was 78.57\%, while RPLIF achieved 78.96\% under the same number of timesteps. When data augmentation was applied, RPLIF achieved an impressive accuracy of 82.40\%, surpassing the previous highest accuracy of 80.7\% by almost 2\% with fewer timesteps establishing a new state-of-the-art result for the CIFAR10-DVS dataset with just 8 timesteps.

\textbf{N-CALTECH101}: Similar to CIFAR10-DVS, the original N-Caltech101 (240×180) was downsampled to 48×48. The VGG9-like network proposed in \cite{li2022neuromorphic} was adopted, and no data augmentation was applied to our experiments on this dataset. The experimental results are shown in the Table \ref{ncaltech101}. Among previous works, RM SNN (Yao et al., 2023) and STCA-SNN (Wu et al., 2023) are notable for surpassing the 80\% accuracy, achieving 81.2\% and 80.88\%, respectively. However, our method, with fewer timesteps, established a new state-of-the-art by achieving an accuracy of 83.35\%, outperforming the above works by 2.15\% and 2.47\%, respectively, which represents a significant improvement in the SNN field.

\textbf{DVS128}: For this dataset, we utilized the same network as in the CIFAR10-DVS experiment, except that max pooling was used instead of average pooling, and the final average pooling layer was replaced with a global average pooling layer \cite{ding2024shrinking}. We also utilized the PLIFNet employed by previous models. Since most previous studies conducted experiments with longer timesteps, we reproduced some prior works to ensure a fair comparison. To preserve the critical details of dynamic gestures, no downsampling was applied. Instead, we directly processed inputs at their original resolution of 128×128. No data augmentation was employed in our experiments on this dataset. As shown in Table \ref{dvs128}, with 8 timesteps, RPLIF achieved the highest accuracy of 95.49\% and 97.22\% with PLIFNet and VGG9 network respectively. Notably, when using the VGG9 network, compared to SSNN, which utilized the same network structure, we achieved an improvement of 2.31\%, greatly enhancing the performance of SNNs under low timesteps on the DVS128 dataset.

\subsection{Additive vs. Multiplicative Dynamics}

The original method increased the threshold by multiplying it with \( \alpha \), which exponentially enhances refractory effects. To investigate whether a simpler additive approach could achieve better performance, we explored threshold adjustment using an additive \( \beta \) under four conditions: 0.4, 0.5, 0.6 and 0.7. Specifically, the threshold was adjusted additively as:
\begin{equation}
    V_{\text{th}} = V_{\text{init\_th}}(1 - S[t]) + (V_{\text{th}} + \beta V_{\text{init\_th}}) S[t]
\end{equation}
where the values of \( \beta \) were selected to maintain comparable ranges of threshold modulation to the original multiplicative approach.

\begin{figure}[tb]
\centering
\includegraphics[
        width=0.475\textwidth,        % 或者指定比例
        trim=15 280 10 270, 
        clip
    ]{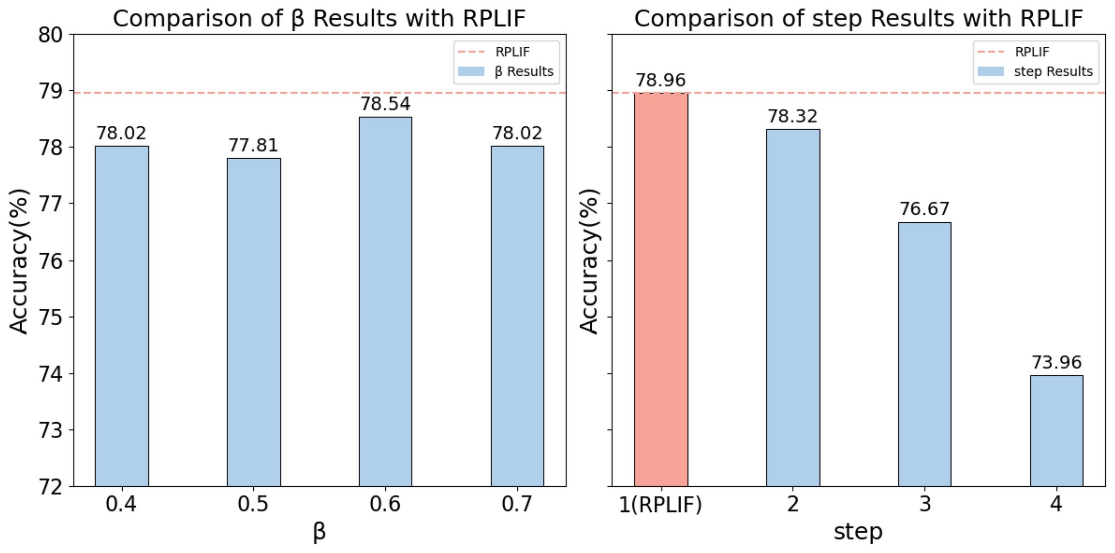}
\caption{Comparison with Additive Threshold Dynamics(left) and Multi-step Refractory Period(right).}
\Description{The bar chart illustrates the accuracy of RPLIF under different settings. The red bars represent the accuracy achieved with the settings adopted in this study, while the blue bars correspond to other settings.}
\label{figure6}
\end{figure}

Fig.~\ref{figure6}(left) shows the performance across different \( \beta \) values. Compared to the multiplicative approach, the additive method consistently yields lower accuracy. We attribute this result to the inherent advantage of multiplicative threshold dynamics, which adjusts the threshold proportionally to its current value. This means that when the threshold is high at the time of spike firing, the increase is correspondingly larger, making the adjustment more adaptive and biologically plausible. In contrast, the additive method applies a fixed increase, which may overlook the neuron's current state and lead to suboptimal refractory effects.

\subsection{Multi-step Refractory Period}

We tested different step values to extend the refractory period. The value of step determines how many timesteps the refractory period lasts. As shown in Fig.~\ref{figure6}(right), when the step was set to 1 (the step adopted in this study), the performance was optimal. However, as the step value increased, the accuracy gradually decreased, suggesting that extending the refractory period had a negative impact on the model’s performance. This could be due to the threshold remaining elevated for an extended period, preventing neurons from firing spikes and transmitting information effectively within a limited number of timesteps.

\subsection{Robustness}
\begin{figure}[tb]
\centering

\includegraphics[
        width=0.475\textwidth,        % 或者指定比例
        trim=0 280 0 270, 
        clip
    ]{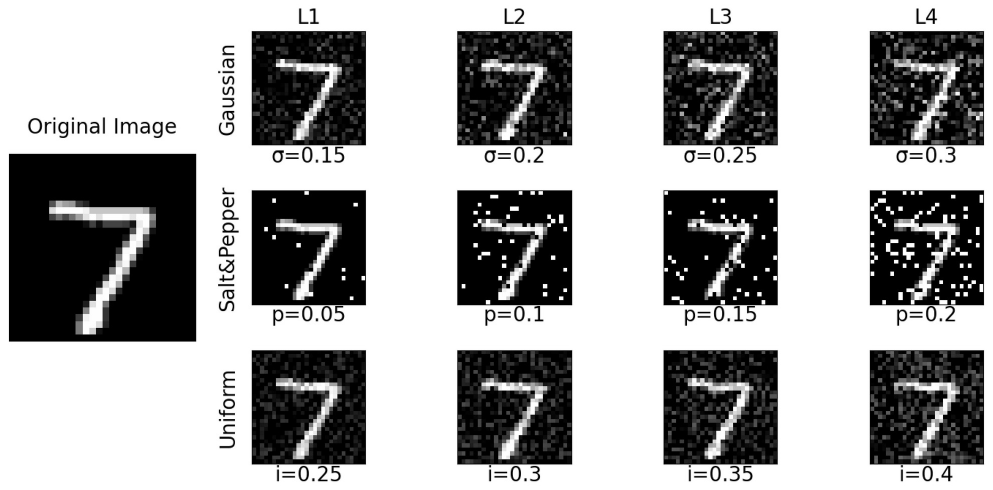}
\caption{Visualization of the original image alongside those affected by four noise intensity levels under three types of noise: Gaussian, salt-and-pepper, and uniform.}
\Description{Visualization of four noise intensity levels under three types of noise: Gaussian, salt-and-pepper, and uniform.}
\label{noise}
\end{figure}

To further demonstrate the robustness of RPLIF in handling anomalous inputs, we compared RPLIF and LIF neurons under three types of noise perturbations: Gaussian, salt-and-pepper and uniform noise. Due to the unique characteristics of neuromorphic datasets, traditional noise augmentation methods are difficult to apply directly. Therefore, following ~\cite{cheng2020lisnn,xu2023constructing}, we chose to validate the robustness of our method on the static MNIST dataset. We adopted the architecture proposed in \cite{fang2021incorporating} (without the voting layer) and tested four different noise levels for each type of noise, ranging from low to high. In Fig.~\ref{noise}, we visualize the image distortion caused by four levels of noise intensity for each type of noise and also provide the specific parameter settings corresponding to each noise level. Specifically, $\sigma$ represents the noise standard deviation in Gaussian noise, p denotes the proportion of noisy pixels in salt-and-pepper noise, and i indicates the noise intensity range in uniform noise. In all cases, larger values correspond to more severe noise interference.

\begin{figure}[tb]
\centering
\includegraphics[
        width=0.475\textwidth,        % 或者指定比例
        trim=10 330 10 330, 
        clip
    ]{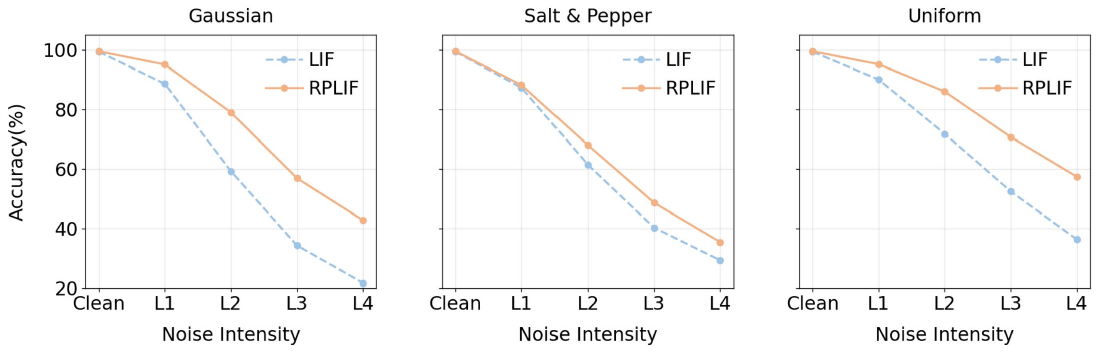}
\caption{The accuracy of RPLIF and LIF on the MNIST dataset under three types of noise—Gaussian, salt-and-pepper, and uniform. "Clean" represents the accuracy without noise, while L1, L2, L3, and L4 correspond to four increasing noise levels in Fig.~\ref{noise}. RPLIF demonstrates superior resilience to anomalous data, highlighting its robustness.}
\Description{The three figures illustrate the accuracy of RPLIF and LIF under Gaussian, salt-and-pepper, and uniform noise conditions. The orange line represents RPLIF, while the blue line represents LIF.}
\label{robustness}
\end{figure}

As shown in  Fig.~\ref{robustness}, in the absence of noise (clean), both neurons achieved over 99\% accuracy on MNIST, given the dataset's simplicity, with RPLIF slightly outperforming LIF. However, as the noise intensity increased, the robustness of RPLIF against anomalous inputs became more evident. Across all three noise types and at every noise level, RPLIF consistently outperformed LIF in accuracy. Notably, under high-intensity noise conditions, RPLIF demonstrated a significant advantage, achieving nearly 20\% higher accuracy than LIF under Gaussian and uniform noise and approximately 10\% higher under salt-and-pepper noise. This highlights the superior robustness of RPLIF, which benefits from the incorporation of the refractory period, enhancing its resistance to noise interference.

\begin{figure}[t]
\centering
\includegraphics[
        width=0.475\textwidth,        % 或者指定比例
        trim=0 250 0 250, 
        clip
    ]{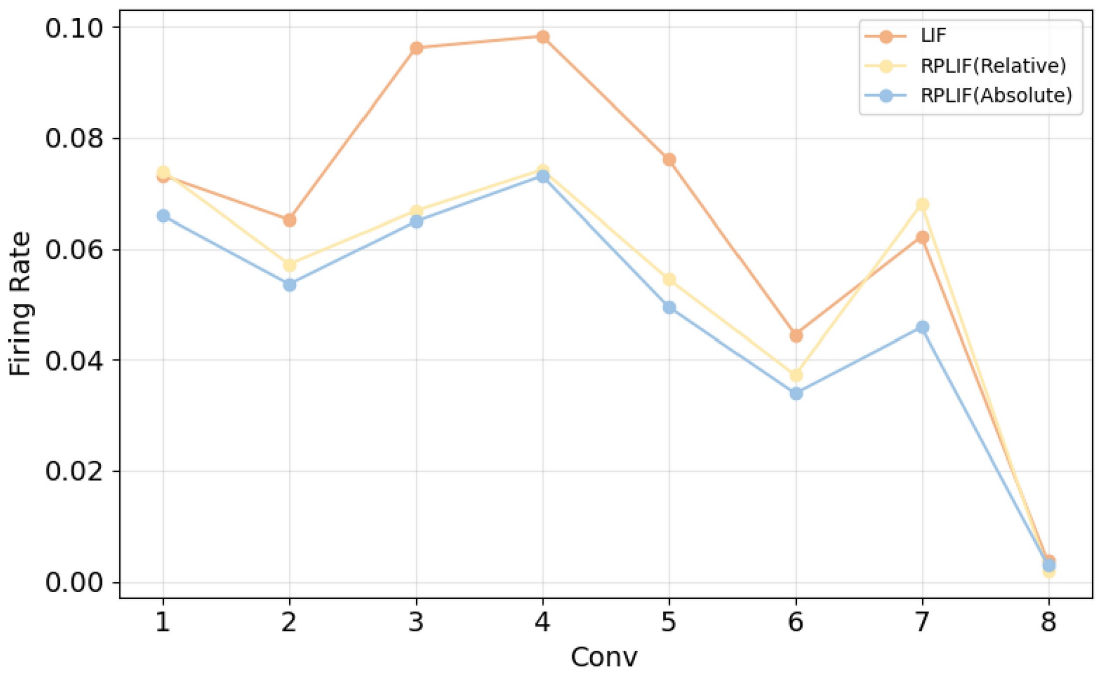}
\caption{The average firing rates of LIF and RPLIF in each of the eight convolutional layers of the VGG network on CIFAR10-DVS. RPLIF exhibits a lower firing rate, which is key to energy efficiency.}
\Description{The orange, yellow, and blue lines represent the firing rates of LIF, RPLIF (relative), and RPLIF (absolute), respectively.}
\label{efficiency_1}
\end{figure}

\subsection{Efficiency}
In SNNs, neurons can only emit spikes when their membrane potential exceeds the threshold. This sparse spiking activity is a key factor contributing to the energy efficiency of SNNs compared to ANNs. The firing rate refers to the average rate or frequency at which a neuron emits spikes over a given period. A higher firing rate results in more spikes per unit time, leading to increased energy consumption.
To evaluate energy efficiency, we computed the average firing rate of all neurons in each of the eight convolutional layers of the VGG network used for CIFAR10-DVS in both RPLIF and LIF models. As shown in Fig.~\ref{efficiency_1}, RPLIF exhibits a lower firing rate across most layers. This difference is particularly pronounced in the intermediate convolutional layers, where RPLIF's firing rate is nearly 30\% lower than that of LIF neurons. This reduction in spike activity is a critical factor in enhancing energy efficiency.

\section{Conclusion}

In this paper, we identify that mainstream neurons in current SNNs overlook the biological refractory period, leading to oversensitivity to continuous and anomalous inputs. To address this, we propose RPLIF, a simple yet effective method that incorporates the refractory period into LIF neurons by dynamically increasing the threshold after each spike. Furthermore, we decoupled the biologically continuous absolute and relative refractory periods into separate threshold dynamics, enabling effective integration into low-timestep SNNs. Our approach effectively mitigates over-excitation and enhances resistance to abnormal signals with minimal computational overhead. 
Experimental results show that our method achieves state-of-the-art accuracy on neuromorphic datasets with low latency.
% Extensive experiments validate the effectiveness of our method. RPLIF achieves state-of-the-art accuracy on CIFAR10-DVS (82.40\%) and N-Caltech101 (83.35\%). Furthermore, it demonstrates the best accuracy reported in DVS128 (97.22\%) under low-latency conditions. 
Further exploratory experiments have also validated the robustness and efficiency of RPLIF.

\begin{acks}
This work is supported by National Natural Science Foundation of China (NSFC 62371325), Sichuan Science and Technology Program 2025NSFJQ0050, Sichuan Science and Technology Program 2024ZDZX0018, and Key Lab of Internet Natural Language Processing of Sichuan Provincial Education Department(No.INLP202402)
\end{acks}

%%
%% The next two lines define the bibliography style to be used, and
%% the bibliography file.
\bibliographystyle{ACM-Reference-Format}
\bibliography{sample-base}
%%
%% If your work has an appendix, this is the place to put it.
\end{document}